\documentclass[conference]{IEEEtran} 
\usepackage{graphicx}
\usepackage{booktabs}
\usepackage{amsmath}
\usepackage{microtype}
\usepackage{inconsolata}
\usepackage{xcolor}

\title{PU-Lie: Lightweight Deception Detection in Imbalanced Diplomatic Dialogues via Positive-Unlabeled Learning}

\author{
  Bhavinkumar Vinodbhai Kuwar\textsuperscript{1,*}, 
  Bikrant Bikram Pratap Maurya\textsuperscript{1,*}, 
  Priyanshu Gupta\textsuperscript{1,*},
  Nitin Choudhury\textsuperscript{1}\\
  
  \textsuperscript{1}Indraprastha Institute of Information Technology Delhi \\
  New Delhi, India \\
  Emails: \{bhavinkumar24212, bikrant24116, priyanshu24130, nitinc\}@iiitd.ac.in \\

  \textsuperscript{*}Contributed equally as first authors.
}

\date{July 2025}

\begin{document}

\maketitle

\begin{abstract}
Detecting deception in strategic dialogues is a complex and high-stakes task due to the subtlety of language and extreme class imbalance between deceptive and truthful communications. In this work, we revisit deception detection in the Diplomacy dataset, where less than 5\% of messages are labeled deceptive. We introduce a lightweight yet effective model combining frozen BERT embeddings, interpretable linguistic and game-specific features, and a Positive-Unlabeled (PU) learning objective. Unlike traditional binary classifiers, PU-Lie is tailored for situations where only a small portion of deceptive messages are labeled, and the majority are unlabeled. Our model achieves a new best macro F1 of 0.60 while reducing trainable parameters by over 650$\times$. Through comprehensive evaluations and ablation studies across seven models, we demonstrate the value of PU learning, linguistic interpretability, and speaker-aware representations. Notably, we emphasize that in this problem setting, accurately detecting deception is more critical than identifying truthful messages. This priority guides our choice of PU learning, which explicitly models the rare but vital deceptive class.
\end{abstract}
\section{Introduction}

Deception plays a pivotal role in strategic communication, shaping the outcomes of negotiations, military strategies, and political campaigns. In environments like the game of Diplomacy, where trust and betrayal are in constant tension, the ability to detect lies can be the difference between victory and defeat. Automated deception detection in such scenarios remains challenging due to the nuanced nature of human language and the extreme imbalance in class distribution.

Deception detection has long been a critical research challenge in computational linguistics, psychology, and security domains. Early efforts primarily focused on manual feature engineering, where linguistic cues such as lexical diversity, hedging, sentiment polarity, and syntactic complexity were analyzed to differentiate deceptive from truthful texts~\cite{hancock2008lying, ott2011finding}. While such methods offered interpretability, they lacked scalability and contextual depth.

With the advent of deep learning, particularly transformer-based models like BERT, deception detection shifted towards end-to-end learning from raw text~\cite{devlin2019bert}. These models achieved substantial improvements across domains such as fake reviews, phishing emails, and online fraud~\cite{ruchansky2017csi, vollmer2021linguistic}. However, most of these approaches presume balanced datasets or moderately skewed distributions, limiting their utility in real-world applications where deceptive instances are sparse and highly contextual.

We focus on the Diplomacy dataset introduced by Peskov et al. (2020), which contains over 17,000 annotated messages exchanged between players. This dataset marked a pivotal shift by offering annotated real-world instances of deception embedded within strategic negotiations. However, only about 4.5\% of these messages are labeled as deceptive, resulting in a highly skewed class distribution that hinders conventional supervised models. Traditional solutions like oversampling, undersampling, or cost-sensitive learning (e.g., class weights or focal loss~\cite{lin2017focal}) have been employed, but these methods often introduce overfitting or rely on noisy/missing negative labels.

To overcome these limitations, an alternative paradigm—Positive-Unlabeled (PU) Learning—has been explored in domains where only positive labels are reliable and negatives are ambiguous. PU learning has shown success in bioinformatics, fraud detection, and recommendation systems~\cite{bekker2020learning, elkan2008learning}, as it estimates the risk of misclassification based solely on the positive class and unlabeled examples. Despite its relevance, PU learning has seen limited application in NLP, and to our knowledge, no prior work has applied it to deception in strategic communication. Our work emphasizes the importance of prioritizing rare deception cases over more abundant truthful ones.

In parallel, models that combine deep learning with structured and interpretable features are gaining traction. Studies like~\cite{rajani2019explain} highlight how integrating human-understandable features (e.g., sentiment, discourse, power dynamics) with neural architectures can boost both accuracy and transparency. In strategic games like Diplomacy, features such as power imbalance, pronoun usage, and game phase offer vital domain-specific context often overlooked by purely neural models. Graph neural networks (GNNs) and dialogue-aware LSTMs have also been explored to model player interactions~\cite{velivckovic2018graph, zhang2021structure}, but these often suffer from overfitting under extreme class imbalance.

Detecting deception is more critical than detecting truthful communication in this setting, as the consequences of missing a lie can significantly affect strategic decisions. Hence, we prioritize models that perform well on the deceptive class. Traditional models often collapse to the majority class, failing to generalize or identify rare but impactful deceptive instances.

To this end, we introduce \textbf{PU-Lie}, a novel application and a compact model for deception detection. It integrates frozen BERT embeddings with handcrafted linguistic and game-specific features and employs a Positive-Unlabeled (PU) learning approach to handle data scarcity and imbalance. PU-Lie not only achieves a new state-of-the-art macro F1-score of 0.60 but does so with only 1,345 trainable parameters---a $650\times$ reduction from prior models.
To this end, we introduce \textbf{PU-Lie}, a novel application and a compact model for deception detection. It integrates frozen BERT embeddings with handcrafted linguistic and game-specific features and employs a Positive-Unlabeled (PU) learning approach to handle data scarcity and imbalance. PU-Lie not only achieves a new state-of-the-art macro F1-score of 0.60 but does so with only 1,345 trainable parameters---a $650\times$ reduction from prior models.

\textbf{Research Gap.} Existing work on deception detection in strategic dialogues largely relies on fully supervised learning, assuming access to balanced datasets with clear negative labels. However, this assumption breaks down in real-world scenarios like Diplomacy, where deceptive messages are rare, unlabeled messages dominate, and truth labels are often ambiguous or missing. Moreover, most prior studies overlook the critical asymmetry in importance—accurately detecting deception is far more impactful than identifying truth. Despite this, few models have explicitly prioritized deception or explored anomaly-based approaches such as Positive-Unlabeled (PU) learning to address this skew. Finally, many existing models are computationally complex, limiting their applicability in low-resource or real-time environments. This highlights the need for a lightweight, interpretable, and deception-focused model that leverages PU learning for effective detection in highly imbalanced dialogue settings.
\textbf{Research Questions:}
This study is motivated by the following core research questions:

\begin{enumerate}
    \item \textbf{Can anomaly detection-inspired techniques such as Positive-Unlabeled (PU) learning effectively address extreme class imbalance in deception detection tasks?}
    \item \textbf{Does reducing the number of trainable parameters by several orders of magnitude (e.g., 650$\times$ fewer) compromise the model’s ability to detect subtle deception cues?}
    \item \textbf{Is a lightweight model, with minimal architecture and handcrafted features, sufficient to capture deceptive intent in complex strategic dialogues?}
    \item \textbf{To what extent do linguistic features—such as hedging, pronoun usage, and sentiment—contribute to identifying lies in human language?}
\end{enumerate}
Our main contributions are:
\begin{itemize}
    \item A PU-learning-based architecture that prioritizes deceptive message detection in highly imbalanced settings.
    \item An interpretable and efficient feature set combining linguistic signals and game-specific metadata.
    \item Thorough experimentation across seven model variants, covering deep, classical, and graph-based paradigms.
    \item Focusing on deceptive messages rather than truthful.
    \item A lightweight model with significantly less trainable parameters, a significant reduced training time and a significant reduced inference time.
\end{itemize}

Our main contributions are:
\begin{itemize}
    \item A PU-learning-based architecture that prioritizes deceptive message detection in highly imbalanced settings.
    \item An interpretable and efficient feature set combining linguistic signals and game-specific metadata.
    \item Thorough experimentation across seven model variants, covering deep, classical, and graph-based paradigms.
    \item Focusing on deceptive messages rather than truthful.
    \item A lightweight model with significantly less trainable parameters, a significant reduced training time and a significant reduced inference time.
\end{itemize}

\section{Related Work}

\textbf{Deception Detection.}Early studies explored linguistic markers such as hedging, sentiment, and syntactic complexity to detect deception \cite{hancock2008lying, ott2011finding, feng2012syntactic}. With the rise of deep learning, models like RNNs and transformers have been applied to deception tasks on domains like fake reviews and misinformation \cite{ruchansky2017csi, vollmer2021linguistic}. However, these assume balanced datasets and often fail under severe class skew.

\textbf{Strategic Dialogue and Diplomacy.} Peskov et al. (2020) introduced the Diplomacy dataset---a rare collection of real deception instances from strategic gameplay. Their BERT + LSTM model set a benchmark but remained resource-heavy and under-optimized for imbalance.

\textbf{Imbalanced Learning.} Traditional methods include class weighting, resampling, SMOTE \cite{chawla2002smote} and focal loss \cite{lin2017focal}, but they still rely on supervised labels for both classes. These assumptions do not hold in deception tasks, where negative labels may be ambiguous or missing.

\textbf{Positive-Unlabeled Learning.} PU learning addresses scenarios with few labeled positives and many unlabeled samples. It estimates true risk using the known positive subset and class prior \cite{bekker2020learning}. Though PU learning is widely used in fraud detection and bioinformatics \cite{bekker2020learning}, it has also been applied to review spam detection in NLP \cite{li2011learning} and classifier learning from limited labels \cite{elkan2008learning}. Our work emphasizes the importance of prioritizing rare deception cases over more abundant truthful ones.

\textbf{Hybrid Architectures.} Models that combine BERT with structured or interpretable features have shown improved generalization and user trust \cite{rajani2019explain}. PU-Lie follows this path by blending semantics with linguistics and domain-specific signals.

\section{Methodology}

We evaluate seven models for deception detection. Our best model, PU-Lie, uses the following components:

\subsection{Dataset}

The Diplomacy dataset consists of 17,289 private messages from 12 complete games. Each instance includes rich game and message metadata. Key fields:

\begin{itemize}
    \item \textbf{messages:} Sequence of text messages in dialogue.
    \item \textbf{speakers, recipients:} Player roles per message.
    \item \textbf{sender\_labels, receiver\_labels:} Deception tags: true (lie), false (truth), NOANNOTATION.
    \item \textbf{game\_score, game\_score\_delta:} Player scores and differences.
    \item \textbf{seasons, years:} Game timeline (e.g., Spring 1902).
    \item \textbf{absolute\_message\_index:} The index the message is in the entire game, across all dialogs
    \item \textbf{relative\_message\_index:} The index of the message in the current dialog

    \item \textbf{game\_id} Unique game.
    \item \textbf{players:} Involved players(Countries).
\end{itemize}

\textbf{Example — Truthful Message:}
\begin{quote}
\textit{Turkey: That sounds good to me. I'll move to defend against Italy while you move west.}
\end{quote}

\textbf{Example — Deceptive Message:}
\begin{quote}
\textit{Italy: We’re friends, right? I believe every single message I’ve sent you all game has been truth...}
\end{quote}

The key challenge lies in identifying deceptive messages (\textasciitilde4.5\%) amid mostly truthful exchanges. We address this by treating unlabeled instances as a mix of positives and negatives under PU learning, focusing our model design on maximizing deceptive class detection performance.

\subsection{Proposed Methodology}
\begin{figure}[htbp]
    \centering
    \includegraphics[width=\linewidth]{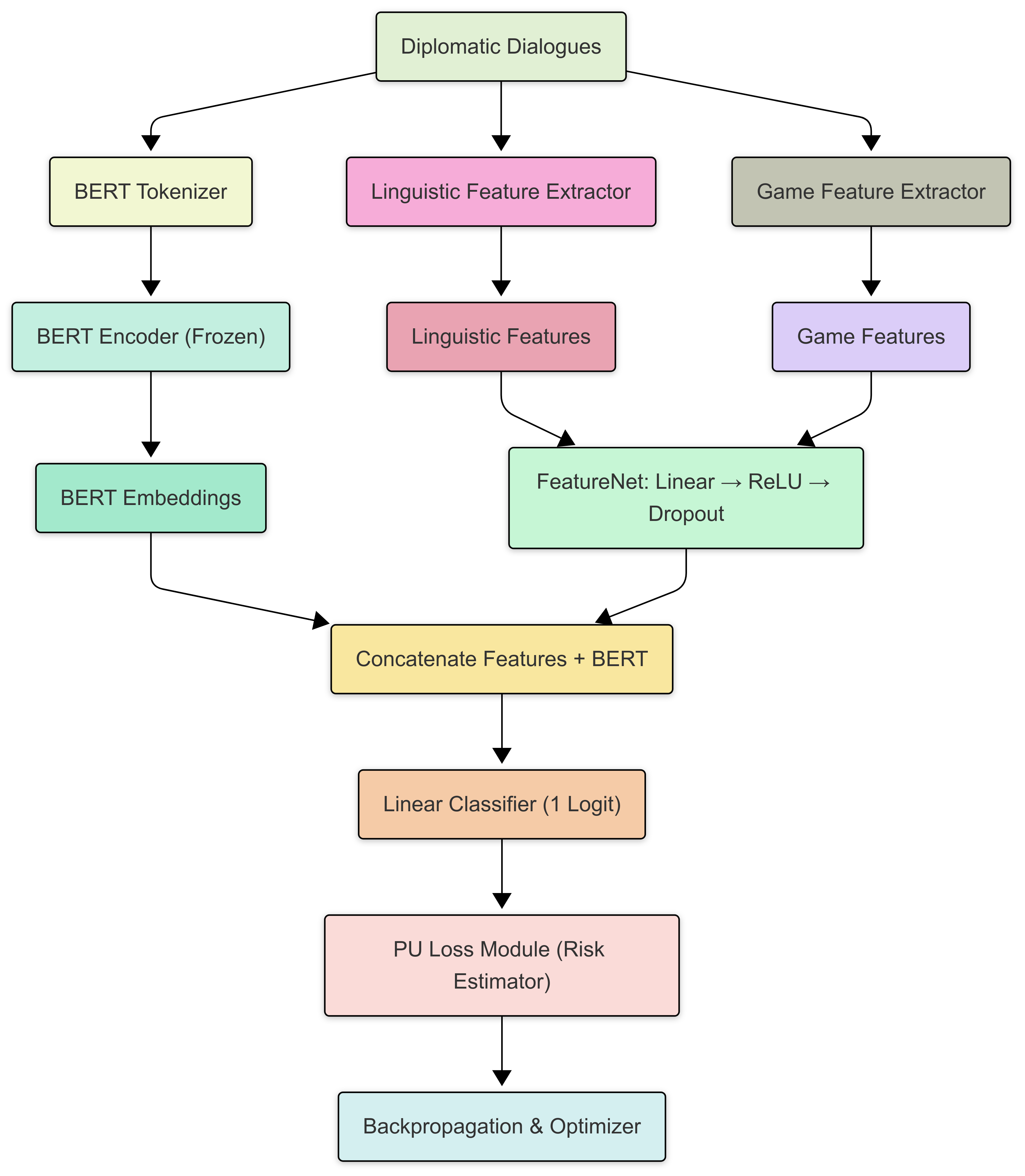}
    \caption{Overview of PU-Lie model architecture combining BERT embeddings, linguistic and game-specific features, and PU learning.}
    \label{fig:flowchart}
\end{figure}

Our proposed system, PU-Lie, is a lightweight and interpretable model designed to detect deception in highly imbalanced diplomatic dialogues. The architecture combines frozen BERT embeddings, handcrafted linguistic and game-specific features, and a PU learning objective to optimize for rare but critical deceptive instances. Figure~\ref{fig:flowchart} illustrates the overall pipeline.

\subsubsection{Input Processing}

All input messages from the Diplomacy dataset are passed through three parallel branches:

BERT Tokenizer and Encoder: Messages are first tokenized and embedded using a frozen \texttt{bert-base-uncased} model to extract contextualized sentence-level representations. These embeddings are fixed during training to reduce model complexity.

Linguistic Feature Extractor: In parallel, a set of handcrafted linguistic features is computed. These include pronoun ratios, hedge word usage, assertiveness scores, and sentiment scores (using VADER), which help interpret deception-related cues.

Game Feature Extractor: Game-related metadata such as season, game score delta, and player roles are extracted to capture strategic context relevant to deception.

\subsubsection{FeatureNet and Fusion}

The linguistic and game-specific features are fed into a small feed-forward subnetwork (FeatureNet), which consists of a linear layer followed by ReLU activation and dropout for regularization. The output of FeatureNet is then concatenated with the frozen BERT embeddings to form a fused representation combining deep semantics with interpretable structure.

\subsubsection{Classification and PU Learning}

The fused representation is passed through a lightweight linear classifier to output a single logit value. Instead of using traditional binary cross-entropy, we employ a PU (Positive-Unlabeled) loss module that estimates the classification risk using the known positive samples (deceptive messages) and the unlabeled set, with a class prior $\pi = 0.05$.

\subsubsection{Optimization}

The model is trained end-to-end using AdamW optimizer with a learning rate of 1e-3 and batch size of 32 for 25 epochs. Only the FeatureNet and classifier layers are updated during training, keeping BERT frozen, resulting in a highly efficient model with only 1,345 trainable parameters.

\section{Experiments}

\begin{table*}[!htb]
\renewcommand{\arraystretch}{1.2}
\caption{Model performance on deception detection. PU-Lie outperforms all baselines while being significantly smaller, faster, and more efficient.}
\label{tab:model_performance}
\centering
\begin{tabular}{@{}lcccccc@{}}
\toprule
\textbf{Model} & \textbf{Macro F1} & \textbf{Trainable Params} & \textbf{Training Time} & \textbf{Inference Time} & \textbf{Epochs} \\
\midrule
TF-IDF + Logistic Regression & 0.3900 & \textbf{1012} & 2 mins & 10 secs & - \\
BERT + Power Embeddings + LSTM & 0.4901 ($\pm$0.01) & 2,300,000 & 18 mins & 30 secs & 10 \\
GNN with Graph Attention & 0.5000 ($\pm$0.01) & 1,101,954 & 22 mins & 40 secs & 10 \\
Oversampled BERT + LSTM & 0.5100 ($\pm$0.01) & 2,300,000 & 19 mins & 40 secs & 10 \\
BERT + Linguistic + Game + LSTM & 0.5453 ($\pm$0.02) & 1,322,626 & 28 mins & 40 secs & 25 \\
BERT + Game Features + LSTM & 0.5471 ($\pm$0.01) & 1,180,994 & 25 mins & 35 secs & 25 \\
\textbf{PU-Lie (Ours)} & \textbf{0.6000 ($\pm$0.01)} & \textbf{1,345} & \textbf{1.7 mins} & \textbf{5 secs} & \textbf{25} \\
\bottomrule
\end{tabular}
\end{table*}

In this section, we conduct comprehensive experiments to evaluate PU-Lie against six alternative models on the Diplomacy deception detection task. We describe our evaluation protocol, model settings, and present detailed analysis of results.

\subsection{Evaluation Protocol}

Given the extreme class imbalance (only 4.5\% deceptive messages), we adopt \textbf{macro F1-score} as our primary evaluation metric. Macro F1 gives equal importance to both deceptive (positive) and truthful (negative) classes, making it more reliable than accuracy under skewed distributions. We also report standard deviation across five runs with different seeds to ensure robustness.

For PU-Lie, we apply \textbf{precision-recall curve-based threshold tuning} to better capture deceptive messages by maximizing recall at an optimal precision.

\subsection{Models Evaluated}

\begin{itemize}
    \item \textbf{TF-IDF + Logistic Regression:} Classical baseline using sparse features. Poor performance due to lack of context \cite{zhou2004learning}.
    \item \textbf{BERT + Game Features + LSTM:} Combines BERT with in-game features and LSTM. Moderate F1 but computationally heavy.
    \item \textbf{BERT + Linguistic + Game + LSTM:} Adds linguistic markers (e.g., pronouns, hedging) to the previous model. Slight F1 boost.
    \item \textbf{BERT + Power Embeddings + LSTM:} Adds strategic power dynamics as embeddings. Performance drops due to feature noise.
    \item \textbf{Oversampled BERT + LSTM:} Uses naive oversampling of positive (deceptive) class. Slight improvement, but overfits.
    \item \textbf{GNN with Graph Attention:} Represents players and messages as a graph. Fails to generalize due to imbalance.
    \item \textbf{PU-Lie (Ours):} Combines frozen BERT, handcrafted features, and PU loss. Best performance with lowest parameter count.
\end{itemize}

\section{Results}


The experimental results, highlighting the comparative performance of all evaluated models, are comprehensively presented in Table~\ref{tab:model_performance}. This table illustrates the macro F1 scores, parameter counts, training time and inferencing time demonstrating the effectiveness and efficiency of the proposed PU-Lie model.

\subsection{Inference from Model Comparison}
\text Although the TF-IDF + Logistic Regression model has slightly fewer trainable parameters than PU-Lie (1,012 vs. 1,345) and both models exhibit comparable training and inference times (approximately 2 minutes and 10 seconds for TF-IDF + LR vs. 1.7 minutes and 5 seconds for PU-Lie), the PU-Lie model significantly outperforms it in terms of macro F1 score (0.60 vs. 0.39). This highlights that while both models are lightweight and efficient, PU-Lie captures deceptive patterns far more effectively due to its integration of contextual BERT embeddings, interpretable linguistic features, and PU learning optimization.

\vspace{0.5em} 

\subsection{Ablation Studies}

\begin{itemize}
\item \textbf{BERT + Game Features + LSTM:} This model uses BERT embeddings, game context features, and a bi-directional LSTM to capture sequential dependencies. It performs better (F1 = 0.5471), showing the value of combining deep language models with game dynamics.

\item \textbf{BERT + Linguistic + Game + LSTM:} Adds linguistic features (e.g., pronouns, hedges) to the previous model. Performance improves marginally (F1 = 0.5453), indicating that LSTMs may not effectively leverage shallow features when trained jointly.

\item \textbf{BERT + Power Embeddings + LSTM:} Introduces handcrafted “power” features reflecting strategic dominance or control, but reduces performance (F1 = 0.4901), suggesting these embeddings may be noisy or redundant.

\item \textbf{Oversampled BERT + LSTM:} Oversamples deceptive messages to balance the dataset before training. Performance (F1 = 0.5100) is better than the power model but still below PU-Lie, indicating that naive balancing does not solve the core problem.

\item \textbf{GNN with Graph Attention:} Models player relationships and game structure as a graph. Despite its sophistication, the model achieves only 0.50 F1, likely due to overfitting on sparse deception data.

\item \textbf{PU-Lie (Ours):} Combines frozen BERT embeddings, linguistic + game features, and PU loss. It achieves the highest macro F1 of 0.60 with only 1,345 parameters, demonstrating superior generalization and efficiency.

\end{itemize}

\section{Discussion}

PU-Lie demonstrates strong performance and efficiency under imbalanced settings. It avoids majority collapse and enhances interpretability.

\textbf{Focus on Deception:} In deception detection, the cost of missing a lie is typically higher than a false alarm. Our use of PU learning reflects this bias by focusing on better identifying deceptive messages and tailoring the architecture for this asymmetric objective.

\textbf{Efficiency and Deployment:} With only 1,345 parameters, PU-Lie is ideal for deployment in edge environments or real-time settings.

\textbf{Limitations:} PU-Lie assumes a static class prior. Cultural nuances or shifts in discourse style may not be captured. Future work may explore dynamic priors and cross-domain transferability.

\section{Conclusion}

We present PU-Lie, a novel, interpretable, and lightweight model for deception detection in imbalanced strategic dialogue. It combines frozen BERT embeddings with handcrafted features and a PU learning framework to outperform heavy baselines using a fraction of the parameters. In addition to its strong performance, PU-Lie is highly efficient, requiring significantly fewer computational resources and enabling faster inference and deployment in low-resource environments. Our results show that explicitly targeting deceptive message detection—rather than treating all messages equally—leads to more robust, interpretable, and generalizable models. Future work includes multilingual extensions, dynamic class prior estimation, and real-time deception detection in interactive settings.

\newpage

\bibliographystyle{IEEEtran}
\bibliography{references}

\end{document}